\documentclass[11pt]{article}
\usepackage{amsmath,graphicx,booktabs,cite}
\usepackage{url}
\usepackage{microtype}
\usepackage{enumitem}
\usepackage[hidelinks]{hyperref}
\usepackage[margin=1in]{geometry}
\setlist{nosep,leftmargin=*}
\newcommand{\irr}{\mathrm{IRR}}
\newcommand{\factf}{\mathrm{FactF1}}
\newcommand{\dcorrect}{D_{\mathrm{correct}}}
\newcommand{\erev}{E_{\mathrm{rev}}}

\title{Recognizing Is Not Reversing: A Controlled Inversion Test of Fact-Preserving News Framing}

\author{
Yi Liu\\
University of Science and Technology of China\\
scnuliuyi@mail.ustc.edu.cn
}

\date{}

\begin{document}
\maketitle

\begin{abstract}
Large language models (LLMs) are increasingly used to analyze and rewrite news, yet current framing studies mainly evaluate generation, detection, or whether rewritten text appears more neutral. They do not directly show whether a model can undo a known framing transformation while keeping the facts fixed. We introduce a controlled inversion test over three established textual realizations of framing: evaluative lexis, agency realization, and information salience. Across 60 news articles and three intervention strengths, this yields 540 paired variants with preserved atomic facts and recorded edits. Across Qwen, DeepSeek, and Kimi, factual preservation remains near 0.84, whereas intervention reversal is 0.044--0.068. Even when both framing type and direction are recognized correctly, pooled reversal reaches 0.071. These results reveal a clear separation between factual fidelity, framing recognition, and framing inversion: recognizing how an article is framed does not imply that the framing can be undone.
\end{abstract}

\noindent
\textbf{keywords:} large language models, news framing, framing inversion, computational social science, robustness

\section{Introduction}
LLMs are becoming practical instruments for computational social science, supporting annotation, explanation, and large scale analysis of social text \cite{ziems2024css,gilardi2023annotation,argyle2023many}. News is a consequential setting because framing can change how the same event is interpreted without requiring a change in its core facts. Communication theory characterizes framing through selection and salience, including the placement of information \cite{entman1993framing}, while news discourse analysis locates framing cues across syntactic, thematic, and rhetorical organization \cite{pan1993framing}. Linguistic work supplies corresponding realization mechanisms: appraisal theory treats evaluative wording as a systematic resource for stance \cite{martin2005appraisal}, and transitivity analyses of news show that grammatical choices about actors and goals shape representations of responsibility \cite{li2010transitivity}. This motivates three realization levels: evaluative lexis, agency realization, and information salience, corresponding to lexical, clause, and discourse resources in news framing.

\begin{figure}[t]
    \centering
    \includegraphics[width=.98\columnwidth]{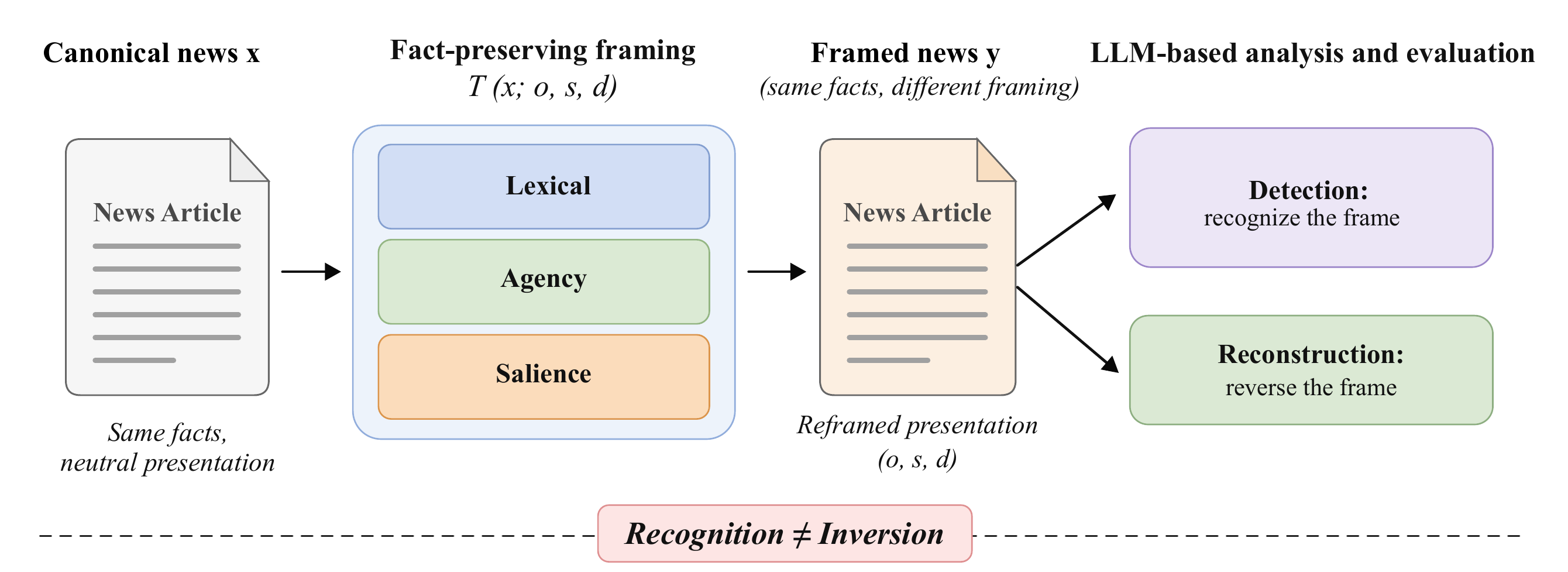}
    \caption{Controlled inversion asks whether a fact-preserving frame can be recognized and then reversed.}
    \label{fig:pipeline}
\end{figure}

\begin{figure*}[!t]
    \centering
    \includegraphics[width=.985\textwidth]{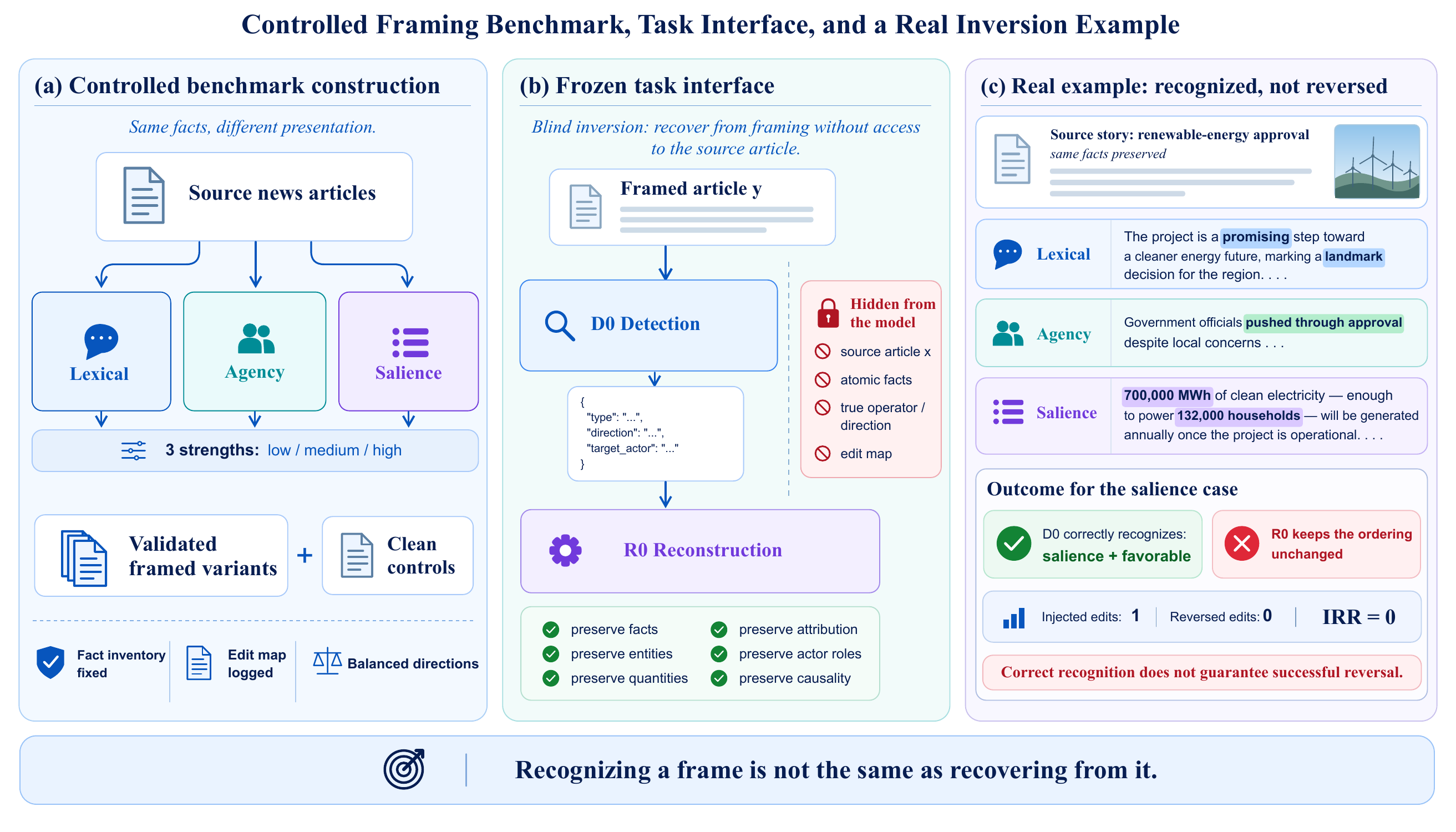}
    \caption{Controlled benchmark, fixed D0 to R0 interface, and a real S011 case. The right panel uses three variants at medium strength from one source to visualize the headline separation: DeepSeek recognizes the favorable salience intervention, while reconstruction preserves the injected ordering ($\mathrm{IRR}=0$).}
    \label{fig:datasetprompt}
\end{figure*}

Recent LLM work measures biased news generation, framing detection, summary bias, and neutralization \cite{yoo2025fair,pastorino2026fifo,pastorino2026decoding,radwan2026unbias,mitra2026breaking}. These tasks reveal whether framing is produced, recognized, or reduced, but not whether a model can reverse a known presentation change. We make the forward transformation observable. From a reference article $x$, we construct $y=T(x;o,s,d)$ while preserving an atomic factual inventory and recording the injected edits. The resulting pair turns framing into an inverse problem: can a model identify what changed in $y$, and can it undo those changes without altering the facts? Figure~\ref{fig:pipeline} summarizes this distinction; Figure~\ref{fig:datasetprompt} gives the full benchmark and task interface.

This formulation makes a distinction that aggregate bias scores are not designed to isolate: recognition and inversion are different operations. The contributions of this paper are summarized as follows:

\begin{itemize}
\item \textbf{Controlled framing inversion benchmark.} We introduce paired news variants in which fact-preserving framing interventions are explicitly constructed and the injected presentation changes are recorded.

\item \textbf{Multi-stage evaluation protocol.} We separate framing recognition, factual preservation, and intervention reversal as independent capabilities, enabling direct measurement of whether recognized framing can be undone.

\item \textbf{Capability separation analysis.} Experiments across three contemporary LLM families reveal that factual fidelity remains substantially easier than presentation-structure recovery, showing that recognizing a framing intervention does not imply successful inversion.
\end{itemize}

\section{Related Work}
\textbf{News framing and computational media bias.}
Classic framing work links news interpretation to selection, salience, responsibility, and recurring issue frames \cite{entman1993framing,scheufele1999framing,semetko2000frames}. Pan and Kosicki operationalize framing through syntactic, script, thematic, and rhetorical structures \cite{pan1993framing}; linguistic accounts complement this view with evaluative stance \cite{martin2005appraisal}, actor representation through transitivity \cite{li2010transitivity}, and the organization of news stories as discourse \cite{bell1991language}. Computational resources then make framing and bias measurable at scale, including the Media Frames Corpus, BASIL, MBIC, and broader surveys of media bias \cite{card2015mfc,fan2019plain,spinde2021mbic,hamborg2019review}.

\textbf{LLM measurement, framing, and rewriting.}
LLMs are now used as annotators and social measurement tools \cite{ziems2024css,gilardi2023annotation,argyle2023many}. News-focused work covers generated bias, framing detection, summary fairness, and neutralization \cite{yoo2025fair,tohidi2025rethink,pastorino2026fifo,pastorino2026decoding,huang2026fairnews,mitra2026breaking}; related reframing work studies span-level bias rewriting and positive reframing \cite{radwan2026unbias,ziems2022reframing}. HELM and broader surveys motivate multi-dimensional evaluation \cite{liang2023helm,gallegos2024survey}, while BBQ, CrowS-Pairs, and BOLD provide complementary bias tests \cite{parrish2022bbq,nangia2020crows,dhamala2021bold}. Prompt formatting and example order affect model behavior \cite{sclar2024prompt,lu2022prompt}, and explicit reasoning can alter performance or faithfulness \cite{wei2022cot,turpin2023unfaithful,lanham2023faithfulness,chochlakis2025cot,zhao2026crowd}. Controlled inversion adds a different target: because the source, factual inventory, and forward intervention are known, reconstruction can be scored against the presentation change that produced the input.

\section{Controlled Framing Inversion}
\subsection{Problem and framing operators}
Let $x$ be a reference news article and $F(x)$ its atomic factual inventory. We construct
\begin{equation}
    y=T(x;o,s,d), \qquad F(y)=F(x),
\end{equation}
where $o$ denotes the framing operator, $s\in\{\text{low, medium, high}\}$ its strength, and $d\in\{\text{favorable, unfavorable}\}$ its direction toward a target actor. The generator also records the injected edit set $E=\{e_1,\ldots,e_k\}$. An evaluated model first predicts the framing, $\hat z=f(y)$, and then reconstructs $\hat x=g(y,\hat z)$. Reconstruction receives the model's own detection output rather than the ground truth intervention.

The operators instantiate the three realization sites motivated above. \textbf{Lexical framing} manipulates evaluative wording, corresponding to linguistic resources for attitude and stance \cite{martin2005appraisal}; sentence and paragraph order remain fixed, and low, medium, and high strength use two, four, and six edits. \textbf{Agency prominence} manipulates grammatical subject, voice, and attribution prominence, drawing on the role of transitivity and actor assignment in news representation \cite{li2010transitivity}; the three strengths use one, two, and three interventions without changing who performed each action. \textbf{Salience framing} manipulates placement and document prominence, directly reflecting framing through selection and salience \cite{entman1993framing,pan1993framing}; it ranges from a local reorder to reorganization of the lead and document order. No operator may add or delete source facts, alter entities or quantities, or introduce new causal claims.

\subsection{Construction and evaluation tasks}
The benchmark contains 60 English news sources, each canonicalized to 180--260 words and represented by 6--10 atomic facts. Three operators at three strengths yield nine framed variants per source, for 540 variants in total; direction is assigned once per source and operator pair with 90 favorable and 90 unfavorable assignments. A disjoint 10-source development set fixes prompt templates and intervention budgets before evaluation.

Controlled variants are produced with GLM-5.2 under a constrained generation and validation procedure. The generator receives the reference article, immutable facts, target actor, operator, direction, and strength, and returns the transformed article together with its edit map. Validation enforces factual preservation, absence of unsupported additions, the requested framing condition, professional plausibility, and operator isolation. All 540 constructed variants satisfy these checks, giving a benchmark of 540 validated framed articles plus 60 clean controls.

Each evaluated model then performs two tasks under a fixed interface. \textbf{Detection (D0)} receives one article, is explicitly told that no presentation problem may be present, considers lexical choice, agency prominence, and salience ordering, and returns a compact JSON record with issue type, direction, target actor, confidence, evidence, and explanation. \textbf{Reconstruction (R0)} receives the framed article $y$ and that same model's D0 record, and produces a factually faithful, minimally framed rewrite while preserving supported propositions, names, dates, quantities, quotations, attribution, actor roles, and causal relations. The source $x$, atomic facts $F(x)$, ground truth operator and direction, and edit set $E$ are withheld; exact original wording is not required.

\textbf{Prompt protocol.} Given documented sensitivity to prompt formatting and order \cite{sclar2024prompt,lu2022prompt}, development fixes task wording, field order, and output schema before evaluation; D0 and R0 are then held fixed across model families. Figure~\ref{fig:datasetprompt} summarizes the benchmark, task interface, and a real example of recognition versus inversion.

\subsection{Metrics}
Detection is summarized by four-class macro-F1, with direction accuracy and the false positive rate on clean news as diagnostics. Reconstruction uses factual preservation and intervention reversal. We compute an atomic fact preservation score, $\factf$, from deterministic lexical and numeric compatibility with the source facts. Let $\erev(\hat{x})\subseteq E$ contain the injected edits reversed in reconstruction $\hat{x}$. We define
\begin{equation}
    \irr(\hat{x};E)=\frac{|\erev(\hat{x})|}{|E|}.
\end{equation}
For conditional analysis, $\dcorrect$ requires the two variables that parameterize the intervention, operator and direction, to match ground truth; target actor text is auxiliary. We pool reversed and injected edits within $\dcorrect$ and its complement. Uncertainty for reconstruction metrics is estimated with 5,000 bootstrap replicates clustered by source article.

\section{Experiments and Results}
\subsection{Evaluation models}
We evaluate DeepSeek-V4-Flash, Qwen-Plus, and Kimi-K2.6 under direct inference with tool and web access disabled; output coverage for each model is reported in Table~\ref{tab:main}. We also evaluate each provider's native thinking configuration on the 180 variants at high strength with the task prompt unchanged, using it as a robustness check across inference configurations; Table~\ref{tab:reasoning} reports matched coverage.

\begin{table}[t]
\centering
\caption{Core direct results. Panel (a) reports four-class macro-F1, direction accuracy, clean false positive rate, joint type and direction recognition, and coverage. Panel (b) reports reconstruction quality with 95\% confidence intervals from bootstrap resampling clustered by source.}
\label{tab:main}
\textit{(a) Detection and calibration}\par\vspace{1pt}
\setlength{\tabcolsep}{1.6pt}
\begin{tabular*}{\columnwidth}{@{\extracolsep{\fill}}lcccccc@{}}
\toprule
Model & F1 & Dir. & FP$\downarrow$ & Exact & D0 $n$ & Clean $n$\\
\midrule
DeepSeek & .247 & .217 & .167 & .128 & 600 & 60\\
Qwen     & .236 & .226 & .250 & .130 & 600 & 58\\
Kimi     & \textbf{.374} & \textbf{.533} & .754 & \textbf{.312} & 571 & 57\\
\bottomrule
\end{tabular*}

\vspace{3pt}
\textit{(b) Reconstruction}\par\vspace{1pt}
\begin{tabular*}{\columnwidth}{@{\extracolsep{\fill}}lccccc@{}}
\toprule
Model & $\factf$ & F1 CI & $\irr$ & IRR CI & R0 $n$\\
\midrule
DeepSeek & \textbf{.840} & [.818,.862] & .044 & [.032,.057] & 540\\
Qwen     & .834 & [.815,.853] & \textbf{.068} & [.051,.085] & 522\\
Kimi     & .837 & [.816,.859] & .054 & [.041,.068] & 513\\
\bottomrule
\end{tabular*}
\end{table}

\subsection{Facts survive while frames persist}
Across all three model families, reconstruction preserves factual content far more reliably than it reverses the injected framing: $\factf$ remains near 0.84, whereas $\irr$ lies at 0.044--0.068 (Table~\ref{tab:main}). This is the paper's first separation: factual fidelity and framing recovery measure different capabilities. The model profiles separate further across stages: Kimi leads Macro-F1 and exact recognition, Qwen attains the highest overall IRR, and DeepSeek the lowest clean false positive rate. Controlled inversion therefore exposes sensitivity, calibration, and recovery as distinct axes rather than a single aggregate rank.

\subsection{Recognizing is not reversing}
Exact recognition occurs in 18.9\% of aligned reconstructions. Conditioning on exact recognition raises pooled IRR from 0.049 to 0.071 (Table~\ref{tab:conditional}), a 45\% relative increase; gains across models range from 0.018 to 0.030. Yet 75 of 1,052 injected edits are reversed in this subset, so 92.9\% of logged interventions remain despite correct type and direction. The S011 case in Figure~\ref{fig:datasetprompt}(c) makes this separation concrete: DeepSeek identifies the salience mechanism and favorable direction, while reconstruction leaves the framed ordering unchanged, yielding $\irr=0$. The conditional counts therefore isolate inversion as an additional operation rather than an automatic consequence of recognition.

\begin{table}[t]
\centering
\caption{Reversal conditioned on recognition. ``Exact'' requires the correct framing type and direction; ``Other'' contains all remaining aligned reconstructions. Pooled counts are 75 of 1,052 reversed edits for Exact and 139 of 2,832 for Other.}
\label{tab:conditional}
\setlength{\tabcolsep}{2.1pt}
\begin{tabular}{lrrrrr}
\toprule
Model & Exact $n$ & Other $n$ & $\irr\mid$Exact & $\irr\mid$Other & Gain\\
\midrule
DeepSeek & 69  & 471  & .066 & .038 & +.027\\
Qwen     & 68  & 454  & .092 & .062 & +.030\\
Kimi     & 160 & 353  & .065 & .046 & +.018\\
Pooled   & 297 & 1278 & .071 & .049 & +.022\\
\bottomrule
\end{tabular}
\end{table}

\subsection{Where framing is encoded matters}
\begin{table}[t]
\centering
\caption{Results by mechanism. Detection reports class F1; reconstruction reports IRR for the three injected framing operators.}
\label{tab:operator}
\setlength{\tabcolsep}{1.7pt}
\begin{tabular}{lcccccc}
\toprule
& \multicolumn{2}{c}{DeepSeek} & \multicolumn{2}{c}{Qwen} & \multicolumn{2}{c}{Kimi}\\
Class & Det. & IRR & Det. & IRR & Det. & IRR\\
\midrule
None     & .212 & --   & .203 & --   & .164 & --\\
Lexical  & .486 & .047 & .393 & .078 & .604 & .041\\
Agency   & .000 & .061 & .163 & .078 & .419 & .091\\
Salience & .290 & .012 & .187 & .025 & .308 & .038\\
\bottomrule
\end{tabular}
\end{table}
The mechanism pattern aligns with the linguistic locus of the intervention. Evaluative lexis is locally explicit and has the highest class F1 for every model. Salience is distributed over placement and document structure and has the lowest reversal rate in all three families. The ranking across stages is especially informative: lexical framing is the easiest mechanism to detect in all three models, yet agency has higher IRR for DeepSeek and Kimi and ties lexical for Qwen. Thus cue visibility does not determine edit recovery. FactF1 remains in the narrow 0.83--0.84 range across operators, concentrating the mechanism effect in framing recovery rather than factual retention.

\begin{figure}[t]
    \centering
    \includegraphics[width=.93\columnwidth]{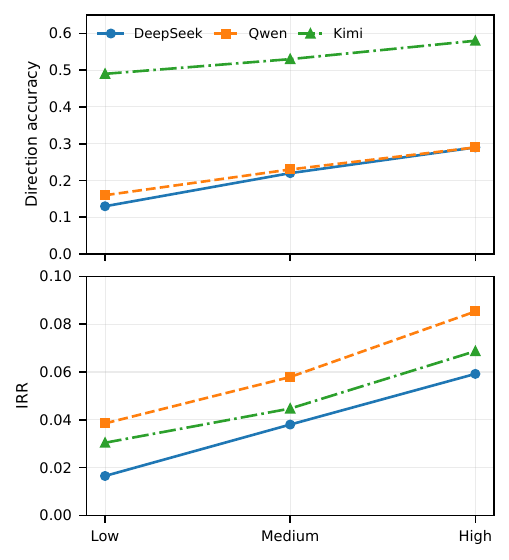}
    \caption{Framing strength acts as a visibility axis: direction accuracy and IRR both increase from low to high strength for all three model families.}
    \label{fig:strength}
\end{figure}

Strength reveals a visibility gradient. From low to high strength, direction accuracy rises by .16, .13, and .09 for DeepSeek, Qwen, and Kimi, while IRR rises by .043, .047, and .038. FactF1 changes much less. The same ordering therefore appears across all three model families and both recognition and inversion: weaker framing preserves the factual substrate while making the presentation change less visible. Importantly, IRR at high strength remains below .09 for every model (.059, .085, and .069), so the inversion separation persists even at the most visible end of the controlled scale.

\subsection{The separation persists under native thinking}
\begin{table}[t]
\centering
\caption{Native thinking on framing at high strength. Deltas are thinking minus direct scores on paired outputs; task prompts are unchanged.}
\label{tab:reasoning}
\setlength{\tabcolsep}{0.55pt}
\begin{tabular}{lrrrrrr}
\toprule
Model & Det. $n$ & R0 $n$ & $\Delta$Dir. & $\Delta$F1 & $\Delta\factf$ & $\Delta\irr$\\
\midrule
DeepSeek & 180 & 180 & -.033 & -.041 & -.007 & +.025\\
Qwen     & 179 & 174 & -.060 & -.042 & -.005 & -.021\\
Kimi     & 157 & 156 & -.286 & -.186 & -.007 & +.000\\
\bottomrule
\end{tabular}
\end{table}
Native thinking leaves the central separation intact. IRR shifts heterogeneously across models (+.025 for DeepSeek, -.021 for Qwen, and +.000 for Kimi), while $\factf$ changes by less than .01 for every model. The separation between recognition and inversion therefore remains visible under a second inference configuration.

\subsection{Regularities across models}
Three regularities recur across all model families: exact recognition raises IRR, salience is the least reversible operator, and increasing intervention strength raises both direction accuracy and IRR while FactF1 remains comparatively stable. The mechanism margins are sizable: lexical detection exceeds the strongest alternative framing class by .185--.206 F1, while salience IRR is 59--80\% below each model's strongest operator. At the same time, the model leading each axis changes: Kimi on recognition, Qwen on reversal, and DeepSeek on clean calibration. The benchmark therefore reveals a reproducible structure across attitude, responsibility, and discourse prominence while keeping recognition, calibration, and recovery empirically distinct.

\section{Discussion}
Controlled inversion turns LLM news analysis into a measurement problem with multiple axes. Clean controls expose calibration, framed variants expose sensitivity, and the source article together with the logged edit map separates factual fidelity from actual correction. The resulting profiles show why corpus coding, editorial auditing, and data curation benefit from reporting these quantities separately rather than compressing them into one quality score \cite{liang2023helm,gallegos2024survey}. They also motivate model selection by stage because recognition, clean calibration, and inversion peak in different model families.

The ranking itself demonstrates why the stages should remain separate. Kimi's exact recognition (.312) exceeds Qwen's (.130), while Qwen has higher IRR (.068 versus .054). A single aggregate score would hide this reversal in model order. Controlled inversion instead makes the handoff from detector to reconstructor measurable under one inverse objective with fixed facts, allowing an auditable pipeline to choose detection and recovery components for the capability each stage actually requires.

For editing with LLMs, the same decomposition also provides a verifiable workflow: detection identifies the presentation mechanism, reconstruction performs the edit, and IRR checks the result against the known intervention rather than against surface fluency alone. This makes controlled inversion useful not only as a benchmark, but also as a design pattern for auditing whether an automated editorial correction actually changed the intended representational choice while preserving content.

The linguistic grounding also yields a direct robustness target. Lexical evaluation is comparatively visible, whereas agency and especially salience require tracking who is foregrounded and where otherwise valid facts are placed. Because paired articles share the same factual inventory, supervision can target the logged presentation change itself: evaluative markers for lexical framing, actor and attribution realization for agency, and information order for salience. Controlled inversion thus turns a broad debiasing goal into verifiable, linguistically grounded corrections while keeping content fixed.

\section{Conclusion}
We introduced controlled inversion for fact-preserving news framing, grounding interventions in evaluative lexis, agency realization, and discourse salience. Across three LLM families, the benchmark separates factual fidelity, framing recognition, and framing inversion, making framing recovery directly measurable. Exact recognition consistently raises reversal but does not determine it; salience is the least reversible operator, and stronger interventions are easier to recover while factual fidelity remains stable. By scoring reconstruction against the known forward intervention, controlled inversion makes recovery the target by construction rather than neutrality by impression. It therefore provides an objective based on recorded edits for LLMs that preserve news facts while changing their presentation.

\end{document}